# Challenges in architecting fully automated driving; with an emphasis on heavy commercial vehicles

Naveen Mohan *, Martin Törngren *, Viacheslav Izosimov *, Viktor Kaznov †, Per Roos †, Johan Svahn †, Joakim Gustavsson *, Damir Nesic *
Corresponding author email: naveenm@kth.se
*KTH Royal Institute of Technology, Stockholm, Sweden
†Scania CV AB, Södertälje, Sweden

*Abstract*—**Fully automated vehicles will require new functionalities for perception, navigation and decision making – an *Autonomous Driving Intelligence (ADI)*. We consider architectural cases for such functionalities and investigate how they integrate with legacy platforms. The cases range from a robot replacing the driver – with entire reuse of existing vehicle platforms, to a clean-slate design. Focusing on Heavy Commercial Vehicles (HCVs), we assess these cases from the perspectives of business, safety, dependability, verification, and realization.**

The original contributions of this paper are the classification of the architectural cases themselves and the analysis that follows. The analysis reveals that although full reuse of vehicle platforms is appealing, it will require explicitly dealing with the accidental complexity of the legacy platforms, including adding corresponding diagnostics and error handling to the ADI. The current fail-safe design of the platform will also tend to limit availability. Allowing changes to the platforms, will enable more optimized designs and fault-operational behaviour, but will require initial higher development cost and specific emphasis on partitioning and control to limit the influences of safety requirements. For all cases, the design and verification of the *ADI* will pose a grand challenge and relate to the evolution of the regulatory framework including safety standards.

## I. Introduction

While autonomous machines are not new, the current strong momentum for autonomous mass-produced vehicles is expected to have broad socio-technical implications, relating to e.g. technology, safety, traffic efficiency, liability, insurances, user-acceptance, legislation[1], and business models[2]. As a consequence, a multitude of stakeholders are already engaging in these issues; new entrants like Alphabet Inc. and established Original Equipment Manufacturers (OEMs) like Volvo Car Group have demonstrated prototypes of highly automated vehicles [1].

To reason about autonomy, several organizations have been defining *levels of automation,* e.g. referring to the wide span of potential settings for autonomous operation such as in different environments and more or less independent of the driver. The Society for Automotive Engineers (SAE) has led one such effort [2]. Their levels of automation range from L0 with no automation and where the driver is responsible for every situation, to L5[3] (highest), which requires the vehicle to handle any situation without human fallback in 'all driving modes' [2]. These levels refer only to capabilities of the vehicle as a whole and are agnostic to implementation details. Modern vehicles already have features that take over control from a human e.g. emergency braking for heavy vehicles, legally mandated in the European Union [3]. However, these features do not correspond to the L5 level of automation, because they do not contain the ability to take appropriate decisions under 'all driving modes'.

Concerns such as business and safety will also heavily influence the choices of architectural strategies for autonomous HCVs. Modern vehicles already have Advanced Driver Assistance Systems (ADAS), which can be seen as steps towards autonomy. To remain competitive, an OEM must identify how much of their previous work can be reused effectively and in which way. The question of how the function of autonomy should be added to their existing product lines thus becomes essential. *In this paper, we focus on HCVs and their architecting challenges w.r.t. achieving L5 level of automation. We analyze how existing platforms, in use by the OEMs, will be affected by the introduction of autonomy.*

The contributions from, and organization of, the paper are as follows; Section II provides the problem statement. In Section III, as a first contribution, we elaborate five architectural cases that describe a span of representative solutions. As a second contribution, Section IV proceeds with an assessment and comparison of the cases w.r.t. business, safety, dependability, verification and realization concerns. As a third contribution, we provide a follow-up discussion in Section V to identify commonalities, tensions among the perspectives as well as overall pros and cons of the architectural cases. We finish in Section VI by providing conclusions and discussing future work.

---

[1] The Vienna convention currently mandates a human driver in the vehicle [33]. Changes have recently been proposed but not approved.
[2] The authors have attempted to categorize sociotechnical factors of Autonomy [32].

[3] Autonomy and fully automated driving are terms used interchangeably in this paper, referring to the SAE standard of L5 driving automation.

II. THE PROBLEM DEFINITION

*A. Problem statement*

Though the first L4 automated vehicles are expected to reach the market around 2020 [4], as of spring 2016 there are not clear standards for assuring their functional safety. Although not explicitly referring to autonomy, the functional safety standard ISO 26262 [5], represents the State Of The Art for the industry for passenger cars. The next revision of the standard, due in 2018 [6], is expected to apply to HCVs as well. The interpretation of the standard for autonomous operation is currently widely debated [7]. However as it is the most relevant standard of today for the automotive industry, this paper will use ISO26262 as a guideline in judging safety related implications.

To define how an L5 vehicle achieves its capabilities we use the 'Observe, Orient, Decide and Act' (OODA) loop [8] as a reference and distinguish between two major parts of the vehicular system i.e. the ADI and the platform. The *ADI* is defined here to correspond to the 'Orient' and 'Decide' parts of the OODA loop. The 'Observe' and 'Act' parts map to both the platform and the ADI since either can contain sensors, and high-level actuation goals are always under the ADI's control. A *platform* is defined here as the set of *components* that provide Electrical/Electronic (E/E) functionality for a vehicle including (but not limited to) hardware and associated low-level control. Components include physical parts such as ECUs, sensors and actuators; and functional components with which we refer to individual software components and logical aggregation of these components into functions.

A *legacy platform* is a platform that is available to an OEM and is currently in use by the OEM for achieving lower levels of automation. It has evolved within the organization over time due to incremental function growth. Typically tailored to an OEM's specific business model, it has seen heavy investments over time. It eliminates the need for redevelopment of core functions and arguably enables existing OEMs to leverage their domain knowledge more effectively. The legacy platform is likely to have followed the common automotive trends and evolved as a largely federated architecture; this means that while functions may be distributed across multiple ECUs, during the evolution of the platform, there has been a grouping and isolation of related functionality e.g. using subnetworks [9].

Brooks distinguishes between *essential complexity* i.e. complexity that arises from realizing product requirements which cannot be designed out of a system, and *accidental complexity* which is possible to be identified and removed from a system by better design [10]. An example of accidental complexity is the incremental growth within legacy platforms that may have caused multiple dependencies between functions. For an existing OEM, or a new entrant collaborating with one, a key question that needs to be addressed is how to add the ADI in the most economical way, with the minimum of accidental complexity, while ensuring adequate safety and dependability. We thus pose the following question: **What are the implications of the ways that the ADI can be coupled with legacy platforms?**

We address this question by (1) stipulating a representative range of architectural scenarios of how the ADI and vehicle platform can interplay, ranging from a strict add-on (platform is reused entirely as a whole), to a 'clean-slate' design (the legacy platform is entirely unchanged), and (2) assessing the architectural solutions through the perspectives of business, safety, dependability, verification and realization.

The particular set of perspectives selected for this paper was selected based on experiences and investigations of the state of the art [11].

*B. Related work*

To the best of our knowledge, the classification of the interactions between the ADI for L5 vehicles and legacy platforms is novel. In this light, the discussion along chosen perspectives (business, safety, dependability etc.) also becomes an original contribution. There is however, a lot of related work on autonomous systems architectures and on specific perspectives such as safety. We provide a brief review of such work here.

Several publications describe prototype autonomous vehicles including e.g. the architecture of 'Bertha', the Mercedes Benz S-class vehicle [12], the European HAVE-IT project [13], Stanford University's DARPA Urban Challenge entry 'Junior' [14] and the A1 car, which won the 'Autonomous vehicle Competition', organized by Hyundai motors in 2010 [15]. These vehicles are designed by adding an ADI to an existing vehicle, with automation levels corresponding to L2 or L3, meaning that there is always a human available to take over if needed. The presence of a human adds a layer of safety and consequently, nearly all of these publications focus on the structure or design of the ADI and deal with the platform only as a tool to achieve the functionality needed. While it is feasible to do so for a prototype vehicle, it is not for a commercially sold, series produced L5 vehicle since extra functional aspects such as safety and dependability have to be designed into a platform and verified in a way that stays in line with the OEMs business model, acceptable safety praxis and within a reasonable cost.

Methodologies and reference architecture for vehicle control system design have been proposed, e.g. by Gordon et al [16] and Behere and Törngren [17]. These methodologies include recommendations for layering for hierarchical control of various functionalities, and also compare and relate to well-known architecture control system patterns such as the subsumption architecture, NASREM [18] and 4D-RCS [19].

The design of novel architectures has received general attention e.g. for fault-tolerant control systems [20]; an example of an adaption of an existing platform to increase reliability and availability is described in [21], where inherent redundancy and reconfiguration is utilized.

Safety and in particular controllability for autonomous vehicles has received a lot of attention in recent times [22] [23]. The automotive industry commonly now appears to take the standpoint that controllability would be set to its worst level resulting in higher Automotive Safety Integrity Levels (ASILs).

III. REPRESENTATIVE ARCHITECTURAL DESIGN APPROACHES

Due to the broad field that this paper covers, some assumptions and delimitations are used to scope and structure the discussion. Issues that deal with the vehicle as a whole i.e. are common to all the different cases e.g., details on how the ADI is structured or modularized internally, are not explicitly discussed. Similarly, issues dealing with communication and collaboration with vehicle-external entities are out of scope.

Figure 1 shows the idealized control topology of a legacy platform. It is modelled as having hierarchical layers of functions, such that functions at the higher layers directly control functions in the layers beneath them. Arbitration is used by functions at the lower layers to resolve multiple control commands. The lowest control layer, the Single Actuator Control Layer (SACL) controls only a single actuator system. One or more of SACL functions are controlled together by the functions at the Vehicle Motion Control Layer (VMCL). The Coordinated Control Layer (CCL) in turn controls one or more functions of the VMCL or below. The choice of separating the CCL and the VMCL was made to distinguish between the characteristics of the sensory inputs available. The CCL consists of functions like ADAS and has access to sensors that observe the environment e.g. cameras, radars, GPS etc., while the VMCL typically has access only to sensors related to the vehicle itself such as vehicle speed sensors, temperature sensors etc. This representation helps anchor the discussions without getting lost in implementation details and reflects upon the principles of how incremental changes have traditionally been incorporated by OEMs.

As an example, within this representation the Adaptive Cruise Control functionality would exist at the CCL and use sensors that observe the environment, such as radar, as input in addition to sensory data from the lower layers. It uses the services of the Cruise Controller at the VMCL, which in turn uses sensors that observe the vehicle itself, such as wheel speed sensors, and commands SACL layer functions such as the brake and engine systems.

Many solutions are possible for how the ADI can interact with the legacy platform. The metric of reusability is applied to filter out cases of interest from amongst these. The ISO 26262 standard distinguishes between the safety-related activities of a new development and a modification of an existing item [5]; an entirely new development generally has a higher need for safety related activities than a modified component. To limit the extent of the safety related activities and the associated costs on the legacy components because of the ADI's influence; we impose a restriction that a component from a legacy platform can only be reused by the ADI if its new usage has not changed from its initial design specification. To comply with this restriction, the usage of the component by the ADI will only be allowed via the existing interfaces and the ADI will meet both the data and timing requirements of the component set during the component's design. This restriction on the reuse of legacy components also implies that new sensors needed by the ADI are added directly to it, and the legacy platform is unchanged. It is also assumed that any component that is part of the legacy platform is mature, safe and reliable. Using the scope set by the delimitations above, we arrive at five representative architectural cases chosen for assessment based on how much of the legacy platform can be reused. The cases show a gradual decreasing reliance on the platform by the ADI as it interfaces with progressively lower layers, allowing for more of the components in the legacy platform to be changed or removed.

Case 1. Robotic ADI

The ADI in this solution is a robot present in the driver seat that is capable of controlling the vehicle in the same way as a human. It uses the same interfaces as a human such as the accelerator and brake pedals, infotainment system etc. There is no electrical connection between the vehicle and the robot.

Case 2. Electrical ADI interface

This solution is characterized by the E/E interfacing of the ADI to the CCL. The ADI is capable of controlling the vehicle through similar platform functions as available to a driver. The ADI is allowed to access the VMCL layer or below only if the driver had the same access before.

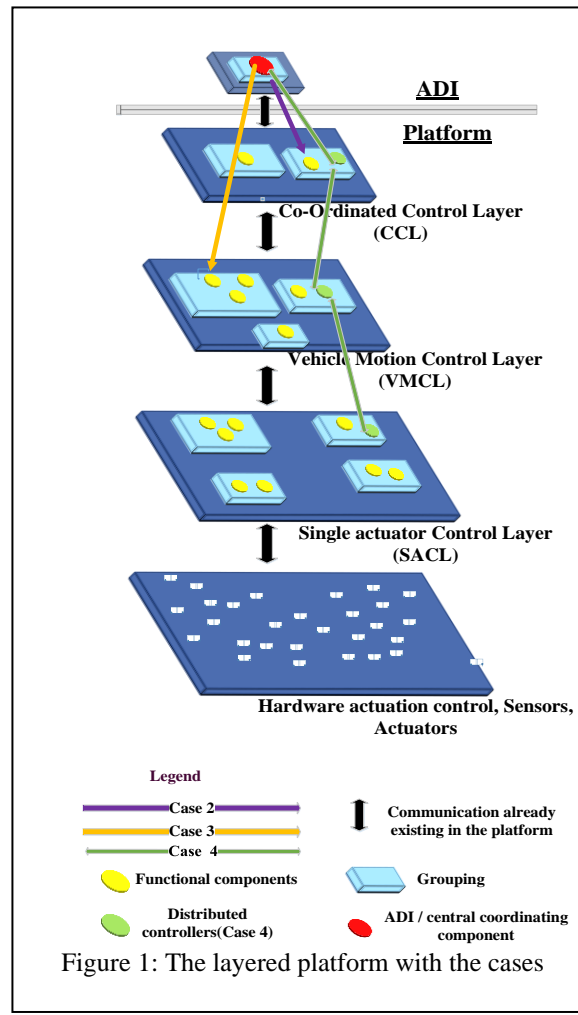

Figure 1: The layered platform with the cases

Case 3. Deep ADI-platform integration

A deeper point of interface into the platform by the ADI characterizes this solution. The ADI has the ability to bypass the CCL layer to access functions at the VMCL, for higher granularity in operation. If needed, CCL functions may be disabled. The ADI cannot control functions at the SACL.

Case 4. Distributed platform ADI

This approach encourages the use of legacy platform components as much as possible but allows for the redefinition or recalibration of the components as needed by the ADI. The ADI can be distributed into the platform. An example of such a redesign would be a hierarchical distributed set of semi-autonomous nodes with a central controlling node at the apex only responsible for goals at a vehicle level. Each tier in the hierarchy, (1) refines the command it receives for the lower tiers or legacy components and (2) filters out information it receives from the lower tiers or legacy components and reports only what is essential for the tiers above.

Case 5. 'Clean-slate design' approach.

This approach deals with the design of the entire vehicle including the platform from scratch such that autonomy is part of the design goals from the beginning. Detailed analysis of this approach will be limited as the absence of the legacy platform also implies that this case could materialize into any of the other cases. It could also lead to unique solutions that are not achievable with the other cases. This approach is used as a delimiting case for the purposes of this paper.

Cases 1 and 5 are extremes in the range of choices as in the former the legacy platform as a whole is used without changes, and in the latter, the platform is designed from scratch, allowing for the most freedom in terms of choosing technologies and components. Cases 2 and 3 differ primarily in the layer at which they interface to the platform and consequently the granularity of control they exert and information they receive from the legacy platform. The architecture of most of the prototypes described in section II.B fall under Cases 2 and 3. The primary goal behind cases 1 to 3 is to use as much of the legacy platform as possible while the primary goals of case 4 and 5 is to have a platform that is optimized for the usage by the ADI. Only Cases 4 and 5 allow modifications to the platform.

IV. COMPARISON OF THE VARIOUS APPROACHES AND SOLUTIONS

A. Business aspects including variability

An HCV is a transport solution tool for either moving people or goods, and is customized to the owners' specific application and their business needs. The need for a high level of application specific customization creates a large amount of variability for all aspects of the vehicle. Frame dimensions and engine power to specific software modules, individual sensors and actuators and more are subject to customization directly affecting all phases of the product lifecycle; and consequently the cost of a vehicle. These commonalities and variabilities are also present in other development artefacts such as requirements, architecture, verification results etc. [24].

Though design for all possible configurations of a common platform represents a very significant undertaking, there is a direct business benefit. HCVs may have a number of different owners and applications during its lifetime and be rebuilt to meet the different owners' unique requirements. 'Re-buildability' increases the resale value of the product purely based on customization abilities provided by the aftermarket [25] and in turn allows the OEM to charge a premium on the first sale. The same thinking also applies to designing platforms where autonomy is an option. The platform should be able to generate both autonomous and non-autonomous variants.

Cases 1 and 2 open up possibilities for third party providers to provide ADIs as independent compatible products, if the interfaces are standardized across organizations. There could then be competition amongst OEM ADIs and third party ADIs. If on the other hand, the interfaces are proprietary, the OEMs will need to design the ADI themselves. The split of the responsibility of liability between the OEM and the ADI vendor, w.r.t. the entire vehicle will be dependent on a combination of who provides the ADI, does the integration and the verification.

Business decisions also have a temporal dimension. A first mover advantage into the market may be considered strategically important, and some extra costs might be acceptable for this purpose. If this were the goal by the OEM, Cases 1 through 3 appear appealing since the existing platforms could be leveraged and the focus would be on the ADI and its platform integration. Needed sensing capabilities could be added to the ADI (raising costs) but in principle enabling maximal platform reuse. It could even be possible to reach this step by outsourcing the autonomy parts to companies from industries that have more experience w.r.t. automation, while the OEM builds up competence with technologies in this field. However, in the long term, Cases 4 and maybe even 5 could be preferred over cases 1 to 3 as long as the specific instantiation of the solution is flexible enough to generate non-automated variants as a subset of its possibilities.

For any vehicle built from a particular platform, an ECU, sensor or software module always has its given physical place. In Cases 4 and 5, the whole cabin could be considered superfluous due to the absence of a human driver, and its removal would greatly reduce the cost of the bill of materials for every autonomous vehicle built. However, to do so, any ECU physically present in the cabin area and necessary for autonomous functionality, would have to move to another location in the vehicle and be designed for a much tougher environment. Removing the cabin would affect the connectivity to the ECUs, the electrical load distribution, and more, implying that such a solution would be very different from the platform of today. Furthermore, these changes within a single modular platform would mean that non-autonomous variants of the vehicle would also be affected by these substantial changes. Thus, there is a cost to achieving autonomy within a single platform.

The other alternative would be to maintain different product lines for autonomous and non-autonomous variants and compromising on 're-buildability'. This could also be used as a technique to decrease platform variability; thereby limiting the extent of safety analysis and verification efforts to specific product lines. There would nevertheless be unavoidable extra costs involved to maintain different product lines.

Which of these alternatives is more preferable, economical and achievable is a crucial business decision for the OEM.

B. Safety

Safety is a key characteristic for the introduction of any potentially disruptive technology; in general, the technology has to be evaluated as being at least as safe as previous solutions, or risk being rejected by legislators, regulators or users. The introduction of autonomy is driven partly by the potential to improve accident statistics; traffic statistics from the U.S.A. show that the mean time between crashes that cause injuries is about 50,000 vehicle hours [26]. For an autonomous vehicle to be acceptable in society, it will likely need to be proven substantially safer than this human driver baseline.

HCVs will have an additional set of specific safety requirements compared to passenger cars, due to a number of reasons:
- Their size implies a risk of causing a large amount of damage to their surroundings.
- They can carry cargo, hazardous material, with implications on safety far beyond their own capabilities.
- They are currently driven by what the public expects to be experts or 'professional drivers'.

Current standards such as IEC 61508 [27] and ISO 26262 [5] which define integrity levels according to risk, would suggest that autonomy related functions should therefore be classified with the highest integrity levels, e.g. with ASIL D in ISO26262. Moreover, there is a risk that existing safety goals may be assigned higher ASILs due to the absence of a human [21] or the changed operational context.

Safety-related dependencies among different functions are critical to keep track of. This includes cataloguing both direct and indirect interactions including e.g. through resource sharing. Non-safety functions should be kept independent of safety critical functions or they will also have to be developed according to the same strict requirements. For all the cases, this means that the platform has to be handled with strict control so that the influence of safety requirements is as limited as possible. The ADI in its role as the decision-making entity (defined by its roles w.r.t. the OODA loop) in the vehicle, will unless care is taken in the architecture (re)design, connect to many functionalities and parts of the vehicle. From this, it can be deduced that the ADI should be modularized in a such a way that highly safety related parts can be separated from less critical parts as much as possible, to limit their influence on the platform. If not controlled, this will cause dependencies and extra associated costs that could ultimately make autonomy unfeasible.

Case 1 provides an interesting option; it promises to leave the platform untouched, also w.r.t. safety measures. However, there are several implications of Case 1. Firstly, to replace the driver, will require additional vehicle behaviour related sensors and diagnostics to provide the inputs a driver would have (sounds, vibrations, smoke etc.). Such sensors are needed for the ADI to take action in case of vehicle errors such as a flat tire (recall that no access to the information from the platform is possible with case 1). Secondly, the ADI will also have limited since the platform is left untouched and since control takes place only through the standard driver interfaces and its corresponding constraints. Thirdly, the OEMs will only be able to partly leverage their extensive domain knowledge (especially related to safety) when the aim is to create an autonomous robot driver rather than an autonomous vehicle.

For Cases 2 or 3 to maintain their strict interfaces, any explicit or implicit interactions between the ADI and functionality part of the legacy platform must be foreseen and handled during integration testing. Case 2 will be the most challenging, as the paths used by the ADI to the actuating components are the least customizable. It will also be necessary for the ADI to have run-time access to a model of the vehicle including its platform, describing various failure modes and symptoms, in order for the ADI to take proper safety measures (such as, if possible, transitioning into a safe state).

The magnitude of the task to classify the functional interactions in detail suggest that ADI should be moved closer to components of the legacy platform instead (e.g. Case 4), to control the possibilities of interference.

For all cases, safe states and safe behaviour of the vehicle must be identified effectively for various operational scenarios. Enabling this in design will impose specific requirements on the architecture of the platforms. If the resulting platforms are not possible to adapt, or if the complexity cannot be managed, it would be required to start over from scratch. Case 5 might thus be better, or even required, to stay competitive in the long run.

*C. Dependability and security*

As a transport tool, a loss of service from an HCV could mean significant economic losses for the owner. Hence, dependability becomes a serious concern. Assuming that in the design of legacy components, reliance was placed on the driver to detect abnormal behaviours, replacing the driver will imply a greater need to detect failures effectively.

With the complexity introduced by autonomy, a resulting decrease in reliability is inevitable. An argument against Case 1 is that the mechanical components of the robot and the additional sensors introduce more parts than compared to Cases 2 to 4.

Current automotive functionalities are typically designed to be fail-safe e.g. active safety functions shut down as soon as a potential fault is discovered. To keep the same reliability and availability as a non-autonomous vehicle, the ADI will instead probably need to be fail-operational for faults in the both the ADI and the platform. To do so, it will need to know not only that a failure has occurred, but also details of the failure itself i.e. specific details from the platform allowing it to exploit the inherent redundancies in the platform, as exemplified in [21]. However, the implicit dependencies (models of the platform in ADI, diagnostics of the platform performed by the ADI) will probably require an increase in the complexity of the ADI-platform interface to enable well-defined autonomous driving, potentially requiring more changes than Cases 1 to 3 allow. Thus while it is nearly impossible to avoid a fail-safe mode of operation w.r.t. errors in the platform as in Cases 1 to 3, Case 4 allows for fail-operational design.

Reliability and availability are usually achieved by means of part quality, fault-tolerance, robustness, and ease of repair. Due to the implications of the loss of service for an HCV, degraded modes have to be precisely identified per use case and fault tolerance becomes a necessity. Operation in the degraded modes has to be then ensured, with appropriate error detection mechanisms and highly reliable design for critical subsystems. Once the system has failed and transitions to degraded mode, the speed and ease of repair (maintainability) becomes essential. This needs a thorough understanding of the in-vehicle E/E systems and excellent diagnostic and troubleshooting capabilities.

None of the present approaches to diagnostics however, to our knowledge, takes the complexity increase due to autonomy into account. Meeting these needs without the ability to change the legacy platform, such as in cases 1 to 3, will be a substantial challenge.

Though this section has focused so far on non-malicious issues, it should be noted that reliability could also be affected by a lack of emphasis on security. While Case 1 is the least affected by the vulnerabilities in the platform, Cases 2 to 4 expose progressively higher attack surfaces (as defined in [28]) to the world as the ADI is susceptible to influence by more elements of the platform.

To summarize, the aspects of dependability are multidimensional, and should be taken into consideration from early in the system design phase. The dependability targets will be challenging to achieve with the additional complexity of autonomy. To effectively address this issue and keep the same dependability levels as the legacy platform, as much flexibility as possible in the redesign of the platform, is desired.

*D. Verification*

Challenges arise for verification from the impact of safety and dependability. Current standards for safety, implicitly point to an increase in the rigor of needed testing e.g. due to higher ASILs. The introduction of redundancy including potentially higher fault tolerance, the essential complexity of autonomy, and the accidental complexity of legacy platforms further increase the scope of what has to be tested with this rigor.

As mentioned in the safety perspective, specific emphasis will have to be placed on integration testing to cover the interactions between the ADI and the platform. Compartmentalization and limiting potential feature interaction will be of special relevance for Cases 2, 3 and 4. Architecture related verification could be simplified for Case 4 and, especially, Case 5 assuming proper architecture design.

Comparing Cases 1 through 3, Case 1 poses a challenge in that the robot driver has to be proven at least as skilled as that of an expert human driver w.r.t. an asynchronously updated platform; this involves dealing with the interactions through the existing mechanical and visual interfaces. Cases 2 and 3 could imply a relatively (compared to case 1) simpler electronic interface for command propagation between the ADI and the platform enabling partial separation of the verification efforts.

For Cases 1 through 3, achieving a fail-safe behaviour by the ADI in all cases, if it is even possible, will require additional verification efforts; there will most likely be an associated, large increase in (a) the number of environments and use cases handled during verification, and (b) the efforts to achieve traceability between system modes and verification evidence. For Cases 4 and 5, in addition, any reconfiguration or fault-tolerance measures will require specific verification efforts.

High safety-related rigor is already quite difficult to achieve within existing functionality in legacy platforms. This is partly due to the lack of a methodology for systematic and efficient verification of such complex systems [29]. It has been frequently suggested that verification techniques have to be updated, e.g. through the adoption of tool supported traceability and formal verification [30]. However, to achieve cost efficiency, formal verification normally needs to be limited to critical parts of the system.

To limit verification activities, architectures that provide composability will be very important. This ultimately points towards Case 5 as the best option. In all cases, modularization of the *ADI*, regardless of its decentralization, will be very important to facilitate the verification of the *ADI*.

*E. Realization*

The architecture of a system is developed based on both functional and extra-functional needs, and is used as an enabler to resolve conflicts between such needs and minimize the accidental complexity of the system. The goals for designing of the architecture are also heavily influenced by the development process and, the structure and culture of the organization.

Automotive organizations have built their E/E systems from the bottom up. Isolation of related functionality from the rest of the system using e.g. subnetworks, has been a traditional way to keep chances of unintended interactions in check. This has also helped in efficient utilization of the limited bandwidth provided by the Controller Area Network (CAN). As such, a large part of the data available to functions on lower layers of the platform is not usually available on the vehicle wide CAN network.

As described in the dependability section, in the absence of possibilities to change the platform (Cases 1 to 3) there will be a need to compensate with supervisory functions to detect failures. Hence, in all the cases, the ADI will need to get information about the state of the vehicle from the platform. Case 1, where informational flow from the platform is not possible, becomes considerably harder to realize. Cases 2 and 3 will also be adversely affected owing to the lack of detail in the information at the CCL and VMCL layers. A theoretic possibility to compensate would be for the ADI to be physically connected to lower layers into any subnetwork or sensors of use, and read the information required. Practically however, this would result in severe cabling and packaging issues as the number of connections increase with increasing need for more detailed information.

There are two issues relating to the needed information flow, (1) the identification of the relevant parameters would need to be performed relatively early during the design, and (2) the bandwidth needed for transmission of this information may stretch the capabilities of the existing network topology. If not tightly controlled, it could lead to a need to redesign of the entire existing vehicle wide networks. It is thus of high importance to control and limit the implications of the information flow between the ADI and the platform. From a realization perspective, it makes sense to instead move the ADI closer to (or into) the platform components to filter and limit this information flow as much as possible as suggested by Case 4.

The dependability and the safety sections highlight the need to modularize the ADI. OEMs have traditionally had an organizational hierarchy that groups resources by their domain expertise. In Cases 1 to 3, from an organizational perspective, the ADI would likely be separate from those who develop the platform. This would lead to some overlap of responsibilities between those developing the ADI via a platform abstraction, and those working directly with the platform and consequently, operational inefficiencies.

Cases 1 to 3 on closer inspection do not seem to provide the clean divide between the ADI and the legacy platform that they were originally intended for. Case 4 follows trends already in place like grouping domain knowledge together in terms of engineering resources, and increasing the reliance put on those who work in the respective domains. Thus, if Case 5 is not an option due to the business aspects such as resource or cost issues, from a technical feasibility perspective, Case 4 appears to be the most promising.

## V. Discussion

In this section, we reflect on the perspectives provided to the architectural cases. We structure the discussion into topics relevant for all cases, topics relating to cases 1-3 which aim at preserving backward compatibility in the platform and minimizing changes, and topics relating to cases 4-5 which take a more pragmatic approach and aim for optimization w.r.t. autonomy.

### A. Similarities and relations between the perspectives.

This paper focusses on HCVs, pointing to specific considerations such as safety and aspects for commercial transport. However, many of the findings from the perspectives, such as the essential complexity of full automation, also apply to passenger cars.

New functionality is worked into existing platforms every day. The difference between incrementally adding functions, e.g. to fulfil a new regulation, and that of autonomy include: (i) the drastic increase in essential complexity (ii) the wide range of socio technical implications that arise due to the potentially disruptive nature of autonomy and (iii) the absence of a driver to deal with unexpected failures.

The safety and verification perspectives recommend the modularization of the ADI to limit the spread of influence of safety critical requirements and rigor of testing respectively. A need to identify the safe states and the degraded modes of the vehicle precisely was noted with the safety and dependability perspectives. Due to the range of scenarios possible, such identification becomes critical to exhaustively verify that the vehicle is able to transition into a safe or a degraded state.

Care must be taken to handle the trade-offs between safety and availability. The ADI needs to provide high availability, at least in the order of what current platforms do and most likely higher, especially for HCVs. Redundancy will be required for safety and also for dependability, e.g. in terms of sensors, diagnostics etc. Designing and verifying such functionality poses several challenges; the ADI has to be robust and reliable while being perceived as safe.

### B. Cases 1-3

To achieve the same capabilities as a skilled human driver to deal with hazardous situations (puncture, fires, …) the ADI will for cases 1 through 3 need specific sensors (sound, vibrations, etc.). A thorough understanding of the platform, including sensing, diagnostics and models, will be crucial in order for the ADI to take appropriate actions. In all these cases, a platform abstraction and interface will be needed.

Cases 1 through 3 will cause higher accidental complexity for several reasons. The platforms have been evolving in a bottom-up fashion for many years, and have many dependencies across the functions and complex arbitration in-between them. Adding the ADI on top will require special care to avoid feature interactions with the existing platform functions. The ADI will moreover be limited by, and have to consider, existing diagnostics and error handling functions in the platform.

Finally, we note that the ADI will not be able to rely on the presence of platform control functions such as cruise control or emergency braking, since these are designed to be fail-silent. Optionally, the ADI could make use of them but must then also be able to detect when they fail and shift to direct brake and throttle control. This could mean that Case 2 is not feasible due to its limitations. Hence, of these three solutions Case 3 appears to be the most viable, albeit from the short-term perspective.

### C. Cases 4-5

The legacy platform is designed for manual driving. It would be unlikely that the same platform was ideal to accommodate autonomy, where no consideration of the human user needs or expectations need to be accounted for in the design.

A strong argument for cases 4 and 5 is the possibility to minimize the accidental complexity. These cases may also be able to provide a stricter separation of functionalities of different criticality, thereby facilitating verification and providing cost-efficient safety. This would however require very careful designs, and a large initial cost. Case 4 is also, in the short term, most likely to increase the variability in the platform and have significant implications for verification and cost.

It must be noted, that even in Case 5, many components at the lowest layers would be reused to save costs. The choices between Case 4 and Case 5 essentially boil down to the problem of what is better for the organization in question; to start with a legacy platform and successively optimize it to achieve autonomy, or to start with a clean-slate and successively add components to achieve autonomy.

Case 4 allows the platform to continue generating non-autonomous variants while evolving towards autonomy and spreads the cost of achieving autonomy over a longer period. Case 5 for an established OEM, involves the maintenance of two different platforms, at least until the new platform is stable and modular enough to support all the non-autonomous variants that its predecessor supported (and probably till the end of the lifetime of the last vehicle of the previous platform).

An interesting question for cases 4 and 5 is what the interface between the compartmentalized ADI and the (new) platform would look like in terms of the information flow.

## VI. Conclusions and Future work

The conclusion arrived at from this paper is that though solutions like Cases 1 to 3 with a strict separation of the ADI appear appealing at first glance; they have a hidden complexity that could make them unsuitable for achieving a functionally safe, fully automated product. A deeper integration between the ADI and the platform, possible only in Cases 4 and 5, will be unavoidable in the long run.

A series of workshops have been planned with the help of 'Innovative Center for Embedded Systems' (ICES) at KTH within the taskforce of 'Autonomous Systems Architectures and Platforms' (ASAP) industrial competence group, to refine and validate the cases, and the findings from the perspectives. Case studies will be conducted at Scania CV and using the Grand Cooperative Driving Challenge [31] competition for validation.

## VII. Acknowledgment

Support from FFI, Vehicle Strategic Research and Innovation and Vinnova through the ARCHER (proj. No. 2014-06260) and FUSE (proj. No. 2013-02650) projects, is acknowledged. Fredrik Asplund from KTH has provided valuable insights for this paper. Sagar Behere from KTH and Magnus Gille from Scania have helped as reviewers.

## VIII. References

[1] Business Insider, [Online]. Available: https://web.archive.org/web/20160218134901/http://www.businessinsider.com/report-10-million-self-driving-cars-will-be-on-the-road-by-2020-2015-5-6?IR=T. [Accessed 18 02 2016].

[2] SAE International, "Taxonomy and Definitions for Terms Related to On-Road Motor Vehicle Automated Driving Systems," Surface Vehicle Information Report, J 3016, no. 01, 2014.

[3] European Parliament, "Regulation (Ec) No 661/2009 Of The European Parliament And Of The Council," 2009.

[4] I.H.S.Automotive, "Emerging Technologies: Autonomous Cars-Not If, But When," IHS Automotive study, http://press.ihs.com/press-release/automotive/self-driving-cars-moving-industrys-drivers-seat, 2014.

[5] International Standards Organization, "ISO 26262:2011; Road vehicles – Functional safety," 2011.

[6] Electronicspecifier.com, [Online]. Available: https://web.archive.org/web/20160218122716/http://www.electronicspecifier.com/events/safety-first-and-second. [Accessed 18 02 2016].

[7] H. Monkhouse, I. Habli and J. McDermid, "The Notion of Controllability in an autonmous vehicle context," in CARS 2015 - Critical Automotive applications: Robustness & Safety, Paris, France, 2015.

[8] J. R. Boyd, "The essence of winning and losing," Unpublished lecture notes; archived at http://pogoarchives.org/m/dni/john_boyd_compendium/essence_of_winning_losing.pdf, 1996.

[9] M. Di Natale and A. Sangiovanni-Vincentelli, "Moving From Federated to Integrated Architectures in Automotive: The Role of Standards, Methods and Tools," Proc. IEEE, vol. 98, no. 4, pp. 603-620, April 2010.

[10] F. J. Brooks, "No Silver Bullet Essence and Accidents of Software Engineering," Computer, vol. 20, no. 4, pp. 10-19, April 1987.

[11] S. Behere, "Reference Architectures for Highly Automated Driving," Doctoral Thesis at KTH Royal Institute of Technology, 2016.

[12] J. Ziegler et al., "Making bertha drive, "an autonomous journey on a historic route"," Intelligent Transportation Systems Magazine, IEEE, vol. 6, no. 2, pp. 8-20, 2014.

[13] R. Hoeger et al., "Highly automated vehicles for intelligent transport: HAVEit approach," in ITS World Congress, NY, USA, 2008.

[14] M. Montemerlo et al., "Junior: The stanford entry in the urban challenge," Journal of field Robotics, vol. 25, no. 9, pp. 569-597, 2008.

[15] K. Chu, J. Kim and M. Sunwoo, "Distributed System Architecture of Autonomous Vehicles and Real-Time Path Planning based on the Curvilinear Coordinate System," in SAE Technical Paper, 2012.

[16] T. Gordon, M. Howell and F. Brandao, "Integrated control methodologies for road vehicles," Vehicle System Dynamics, vol. 40, no. 1-3, pp. 157-190, 2003.

[17] S. Behere and M. Törngren, "A functional architecture for autonomous driving," in Proceedings of the First International Workshop on Automotive Software Architecture, 2015.

[18] J. S. Albus, H. G. McCain and R. Lumia, NASA/NBS standard reference model for telerobot control system architecture (NASREM), National Institute of Standards and Technology Gaithersburg, MD, 1989.

[19] R. Madhavan, Intelligent Vehicle Systems: A 4D/RCS Approach, New York, NY, USA, 2007.

[20] E. Dilger, T. Führer, B. Müller and S. Poledna, "The X-by-wire concept: Time-triggered information exchange and fail silence support by new system services," 1998.

[21] S.Behere, X.Zhang, M.Törngren and V.Izosimov, "A functional brake architecture for autonomous heavy commercial vehicles," Unpublished; Accepted as SAE Technical paper 2016-01-0134, 2016.

[22] CARS Workshop, "Proceedings of Critical Automotive applications: Robustness & Safety," Paris, 2015.

[23] WASA-15, "Proceedings of Workshop on Automotive Software Architectures," Montréal, 2015.

[24] K. Pohl, G. Böckle and F. J. van der Linden, Software Product Line Engineering. Foundations, Principles, and Techniques, Springer-Verlag Berlin Heidelberg, 2005.

[25] G. J. Thuesen and W. Fabrycky, Engineering economy, 9th, Prentice Hall, 2001.

[26] S.E.Shladover, "An Automated Highway System as the Platform for Defining Fault-Tolerant Automotive Architectures and Design Methods," in NSF CPS workshop, 2011.

[27] International Electrotechnical Commission, IEC 61508:2010; Functional Safety of Electrical/Electronic/Programmable Electronic Safety-related Systems, 2010.

[28] J. Petit and S. E. Shladover, "Potential cyberattacks on automated vehicles," Intelligent Transportation Systems, IEEE Transactions on, vol. 16, no. 2, pp. 546-556, 2015.

[29] Bernhard Schätz et al., "Integrated CPS Research Agenda and Recommendations for Action. CyPhERS (FP7-ICT support action, contract no. 611430), project final deliverable," 2015.

[30] J. Cobleigh, G. Avrunin and L. Clarke, "Breaking up is hard to do: An evaluation of automated assume-guarantee reasoning," ACM Transactions on Software Engineering and Methodology, vol. 17, no. 2, pp. 1-52, 2008.

[31] Grand Cooperative Driving Challenge, http://www.gcdc.net/en/, 2016.

[32] ARCHER project, "Autonomy Mindmap," KTH mechatronics, 2016. [Online]. Available: www.kth.se/itm/autonomymindmap. [Accessed 2016].

[33] United Nations, "Vienna Convention on the Law of Treaties, vol. 1155, p. 331," United Nations, Treaty Series, 1969.